\documentclass[letterpaper, 10 pt, conference]{ieeeconf}  

\IEEEoverridecommandlockouts                              

\usepackage{graphics}    
\usepackage{graphicx}    
\usepackage{epsfig}      
\usepackage{amsmath}     
\usepackage{amssymb}     
\usepackage{booktabs}    
\usepackage{multirow}    
\usepackage{cite}        
\usepackage{url}         
\usepackage{xcolor}      
\usepackage{bm}

\title{\LARGE \bf
Wild-Drive: Off-Road Scene Captioning and Path Planning via Robust Multi-modal Routing and Efficient Large Language Model
}

\author{Zihang Wang, Xu Li,~\IEEEmembership{Member,~IEEE}, Benwu Wang, Wenkai Zhu, Xieyuanli Chen,~\IEEEmembership{Member,~IEEE},\\ Dong Kong,~\IEEEmembership{Member,~IEEE}, Kailin Lyu, Yinan Du, Yiming Peng, Haoyang Che
\thanks{This work was supported in part by the Primary Research and Development Plan of Jiangsu Province under Grant BE2022053-5, and in part by the National Natural Science Foundation of China under Grant 62473099. (\textit{Corresponding author: Xu Li.})}
\thanks{Zihang Wang and Benwu Wang contributed equally to this work.}
\thanks{Zihang Wang, Benwu Wang, Wenkai Zhu, Yinan Du, and Yiming Pen are with the School of Instrument Science and Engineering, Southeast University, Nanjing 210096, China (e-mail: wangzihanggg@hotmail.com; 230228939@seu.edu.cn; 230248520@seu.edu.cn; 220243715@seu.edu.cn; 230248519@seu.edu.cn).}
\thanks{Xu Li is with the School of Instrument Science and Engineering, Southeast University, Nanjing 210096, China, and also with the Southeast University Nanjing Jiangbei New Area Innovation Research Institute, Nanjing, China (e-mail: lixu.mail@163.com).}
\thanks{Xieyuanli Chen is with the College of Intelligence Science and Technology, and the National Key Laboratory of Equipment State Sensing and Smart Support, National University of Defense Technology, Changsha 410008, China (e-mail: chenxieyuanli@hotmail.com).}
\thanks{Dong Kong is with the School of Transportation, Shandong University of Science and Technology, Qingdao 266590, China (e-mail: kd.trans@sdust.edu.cn).}
\thanks{Kailin Lyu is with the Institute of Automation, Chinese Academy of Sciences, Beijing 100190, China (e-mail: lvkailin2024@ia.ac.cn).}
\thanks{Haoyang Che is with the Jetour Auto, a Chery Auto company, and also with the Data \& Intelligence Center,   a division of Chery Auto company, Anhui 241000, China (e-mail:
bigdatache@qq.com).}
}

\begin{document}

\maketitle
\thispagestyle{empty}
\pagestyle{empty}

\begin{abstract}

Explainability and transparent decision-making are essential for the safe deployment of autonomous driving systems. Scene captioning summarizes environmental conditions and risk factors in natural language, improving transparency, safety, and human--robot interaction. However, most existing approaches target structured urban scenarios; in off-road environments, they are vulnerable to single-modality degradations caused by rain, fog, snow, and darkness, and they lack a unified framework that jointly models structured scene captioning and path planning. To bridge this gap, we propose Wild-Drive, an efficient framework for off-road scene captioning and path planning. Wild-Drive adopts modern multimodal encoders and introduces a task-conditioned modality-routing bridge, MoRo-Former, to adaptively aggregate reliable information under degraded sensing. It then integrates an efficient large language model (LLM), together with a planning token and a gate recurrent unit (GRU) decoder, to generate structured captions and predict future trajectories. We also build the OR-C2P Benchmark, which covers structured off-road scene captioning and path planning under diverse sensor corruption conditions. Experiments on OR-C2P dataset and a self-collected dataset show that Wild-Drive outperforms prior LLM-based methods and remains more stable under degraded sensing. The code and benchmark will be publicly available at https://github.com/wangzihanggg/Wild-Drive.


\end{abstract}

\section{INTRODUCTION}


Autonomous navigation is a fundamental capability for self-driving vehicles and mobile robots~\cite{in1}, yet its large-scale and safe deployment depends not only on perception and planning accuracy but also on whether decisions are interpretable, traceable, and communicable~\cite{in2, in3}. In off-road and unstructured environments, frequent sensor degradations, environmental uncertainty, and long-tail hazards make black-box planning harder to trust and validate for safety~\cite{in5}. In this context, scene captioning has emerged as a key approach to improving system transparency and safety by summarizing environmental conditions and risk factors in natural language~\cite{in6}. It presents safety-critical cues in a human-understandable form, facilitating natural human--robot interaction and online correction, while serving as an auditable intermediate representation for downstream planning and trustworthy navigation.

\begin{figure}
    \centering
    \includegraphics[width=1\linewidth]{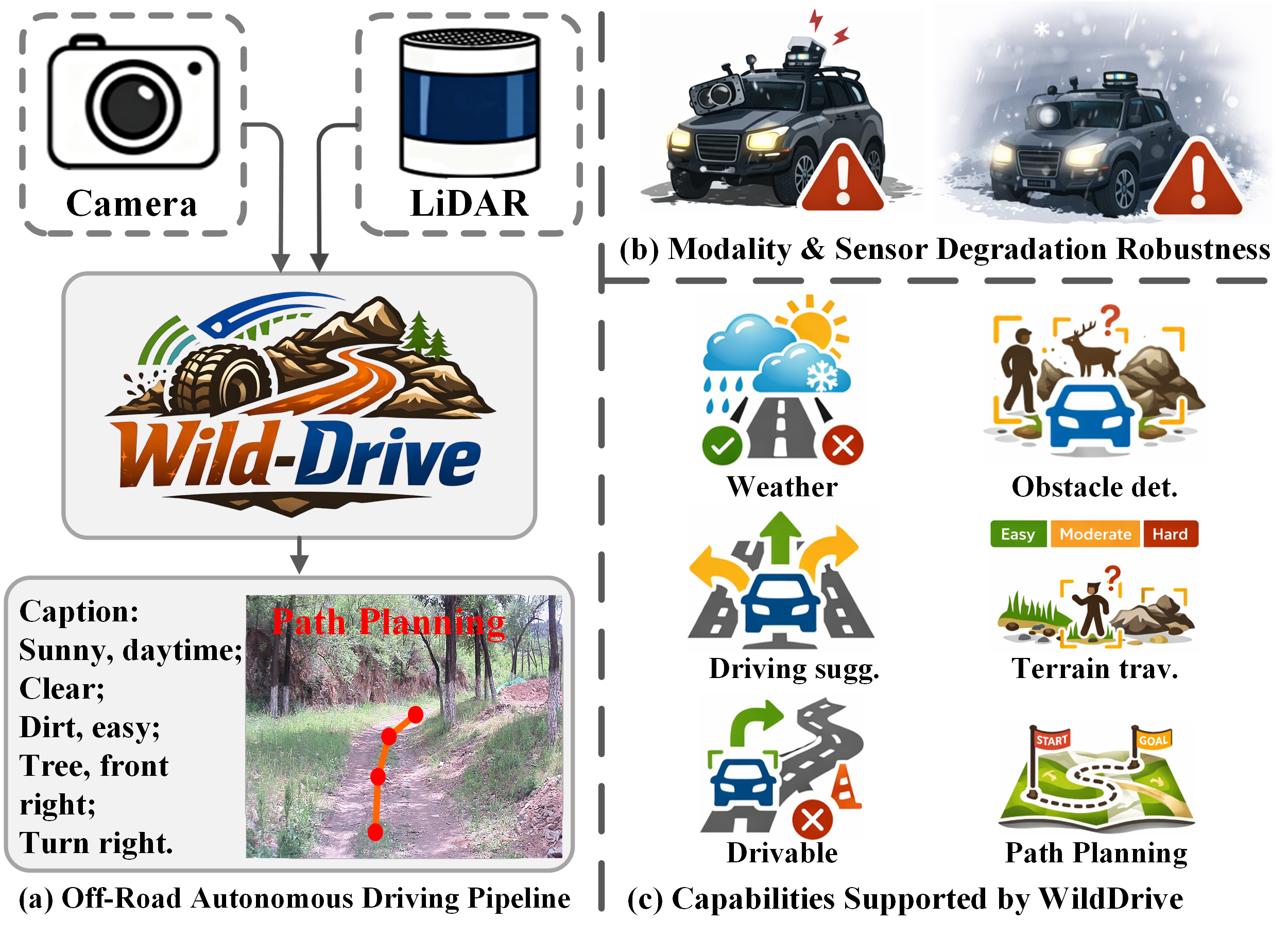}
    \caption{Wild-Drive unifies camera-LiDAR as input for off-road scene structured captioning and path planning, and provides adaptability to sensor corruption.}
    \label{fig:abstract}
\end{figure}


With advances in computer vision and large language models (LLMs)~\cite{in7, in8}, scene understanding and visual question answering for autonomous driving have made significant progress. Most existing works, however, focus on a single sensor modality or structured road settings. For example, Orion~\cite{in9} uses surround-view cameras for scene understanding and planning; LiDAR-LLM~\cite{in10} explores LiDAR-based 3D scene captioning to advance outdoor 3D semantic descriptions; and BEV-LLM~\cite{in11} further integrates camera and LiDAR to handle complex traffic semantics in urban environments. Nevertheless, these methods pay limited attention to off-road autonomy. Compared to urban roads, off-road environments lack clear lanes and rules, and adverse conditions such as darkness often cause single-modality corruption, making fixed fusion strategies less robust. This motivates multimodal approaches that remain reliable for off-road understanding under sensor corruption.



To address these challenges, we propose Wild-Drive, a unified framework for off-road scene captioning and path planning (as shown in Fig.~\ref{fig:abstract}). Wild-Drive is optimized for sensor corruptions and single-modality corruptions commonly encountered in off-road environments. We adopt strong modern backbones~\cite{in7, in12} to extract camera and LiDAR features, and introduce MoRo-Former as a multimodal-to-LLM bridge. Through task-conditioned modality routing and token compression, MoRo-Former adaptively aggregates reliable information under degraded sensing. We then integrate an efficient LLM to generate structured scene captions, and further introduce a planning token together with a gate recurrent unit (GRU) decoder to predict future trajectories, forming an interpretable perception-to-planning pipeline. In addition, to systematically train and evaluate structured outputs, we build the OR-C2P Benchmark, which unifies off-road captioning and planning tasks and covers diverse sensor degradation conditions.

In summary, our main contributions are as follows:
\begin{itemize}
    \item We propose Wild-Drive, a unified framework for off-road scene captioning and path planning, improving robustness and interpretability under single-modality degradation via task-conditioned modality routing and token compression.
    \item We build the OR-C2P benchmark to systematically train and evaluate structured captioning and planning in off-road scenarios, covering diverse sensor corruption conditions. The benchmark will be publicly available.
    \item We conduct extensive experiments on OR-C2P benchmark and a self-collected dataset, showing that Wild-Drive outperforms prior LLM-based methods and remains more stable under degraded sensing.
\end{itemize}

\section{RELATED WORK}

\subsection{Off-Road Autonomous Driving}

Prior work on off-road autonomous driving mainly focuses on drivable-area detection~\cite{rw1} and terrain traversability estimation~\cite{rw3, rw4} in unstructured environments. Recent advances in CNN and Transformer architectures have significantly improved segmentation and traversability prediction under challenging conditions~\cite{rw5, rw6}, where noise-robust designs and attention mechanisms further enhance performance in rain, fog, and low-light scenarios. Meanwhile, multimodal fusion frameworks leverage uncertainty modeling and cross-modal attention to strengthen perception and mapping in complex terrains~\cite{rw8, rw9}. Beyond perception, TopoPath~\cite{rw10} has explored mapless end-to-end off-road navigation, where Transformer-based models predict local trajectories directly from raw point clouds and coarse priors. On the dataset side, RELLIS-3D~\cite{rw11} provides early multimodal data to promote robust perception in natural environments; UniScenes~\cite{rw12} further adopts 3D occupancy as a core representation for unified off-road modeling; and ORAD-3D~\cite{rw13} scales up multimodal off-road data with diverse adverse weather and illumination conditions, offering a more deployment-oriented benchmark for studying sensor degradation and robust multimodal understanding. However, existing studies still lack an LLM-based unified framework that couples interpretable scene captioning with path planning for off-road environments, as well as a systematic benchmark and evaluation protocol tailored to this setting.

\subsection{Scene Caption for Autonomous Driving}


Scene captioning is crucial for interpretable and explainable autonomous driving, as it describes the environment or answers safety-critical queries in natural language, making the model’s focus and decisions more transparent. Early efforts mainly adopt visual question answering (VQA) paradigms~\cite{rw14} and CLIP-based captioning/retrieval~\cite{rw15}. More recent works include ADAPT~\cite{rw16}, which unifies decision reasoning with video description, and BEV-TSR~\cite{rw17}, which performs text retrieval in camera-derived BEV representations. LiDAR-LLM~\cite{in10} explores LiDAR-based 3D scene captioning but is limited by a single modality; BEV-LLM fuses camera and LiDAR for multimodal descriptions; and ${\mathrm{TOD}}^{3}\mathrm{Cap}$~\cite{rw19} advances to object-level 3D dense captioning with 3D boxes and fine-grained texts. However, these methods largely target urban driving and rarely address off-road settings, where rain, fog, snow, and darkness frequently induce single-modality degradation and reduce caption robustness. To bridge this gap, we propose an efficient LLM-based framework for off-road structured captioning and path planning, leveraging modality routing to remain reliable under degraded sensing. We also introduce the OR-C2P Benchmark to provide a unified training and evaluation protocol for structured off-road scene captioning and planning.

\section{PROPOSED METHOD}

\begin{figure*}
    \centering
    \includegraphics[width=1\linewidth]{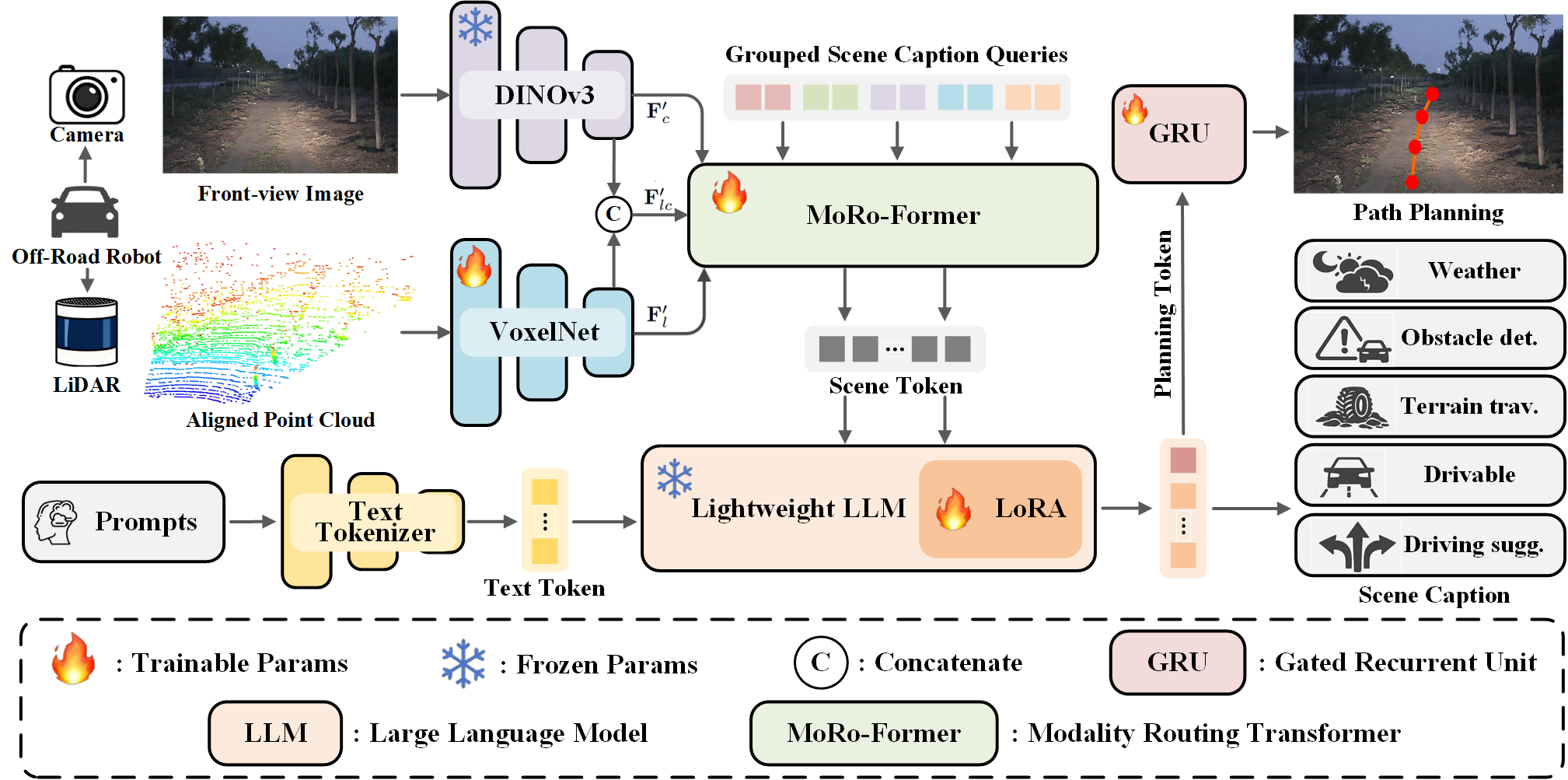}
    \caption{The overview of our proposed Wild-Drive. It fuses camera–LiDAR features into multimodal tokens and uses MoRo-Former for query-conditioned routing and token compression. An LLM generates scene captions and planning tokens, which a GRU-based planner decodes into multimodal trajectories.
}
    \label{fig:method}
\end{figure*}

\subsection{Overall Pipeline}


Fig.~\ref{fig:method}  illustrates the overall pipeline of WILD-Drive. We adopt modern backbones to extract features from the monocular image and LiDAR point cloud, and further construct fused multimodal features. Then, MoRo-Former compresses tokens and performs routing-based aggregation across modalities conditioned on task queries for diverse off-road tasks. Next, an efficient LLM combines the compressed scene tokens with a prompt to generate a scene description. Finally, a generative planner predicts multimodal trajectories by conditioning on planning tokens.

Specifically, given an aligned LiDAR point cloud $\mathbf{P}_l \in \mathbb{R}^{N \times 4}$ and a monocular image $\mathbf{I}_c \in \mathbb{R}^{H \times W \times C}$, we voxelize and encode $\mathbf{P}_l$ using VoxelNet~\cite{in12} to obtain voxel features, which are then projected to the BEV plane and processed by 2D CNNs to produce BEV features
$\mathbf{F}_l \in \mathbb{R}^{H_l \times W_l \times D}$,
where $(H_l, W_l)$ denotes the BEV resolution and $D$ is the channel dimension.
We encode $\mathbf{I}_c$ with DINOv3~\cite{in7} to obtain image features
$\mathbf{F}_c \in \mathbb{R}^{H_c \times W_c \times D}$,
where $(H_c, W_c)$ is the spatial resolution of the camera features.
We flatten them as
$\mathbf{F}'_l \in \mathbb{R}^{(H_lW_l)\times D}$ and
$\mathbf{F}'_c \in \mathbb{R}^{(H_cW_c)\times D}$,
and concatenate to form fused LiDAR--camera tokens
$\mathbf{F}'_{lc} \in \mathbb{R}^{(H_lW_l + H_cW_c)\times D}$.

To handle perception challenges in off-road environments—such as corrupted camera imagery and LiDAR sensor failures, we introduce MoRo-Former as a bridge between the multimodal encoder and the LLM. MoRo-Former decodes a set of off-road task queries
$\mathbf{Q} \in \mathbb{R}^{N_q \times D}$,
using $\mathbf{F}'_l$, $\mathbf{F}'_c$, and $\mathbf{F}'_{lc}$ as keys/values. We employ three expert decoders (LiDAR, camera, and LiDAR--camera fusion), followed by hard routing aggregation and token compression to obtain compact multi-expert scene representations.
The LLM then produces structured, task-specific responses by conditioning on the compressed expert tokens and user instructions. Finally, we map the LLM planning tokens $\mathbf{s}$ to the action latent space $\mathbf{z}$ using two MLP layers, and decode multimodal trajectories from $\mathbf{z}$ with a GRU decoder.

\subsection{Modality Routing Transformer with Grouped Queries}

\begin{figure}
    \centering
    \includegraphics[width=1\linewidth]{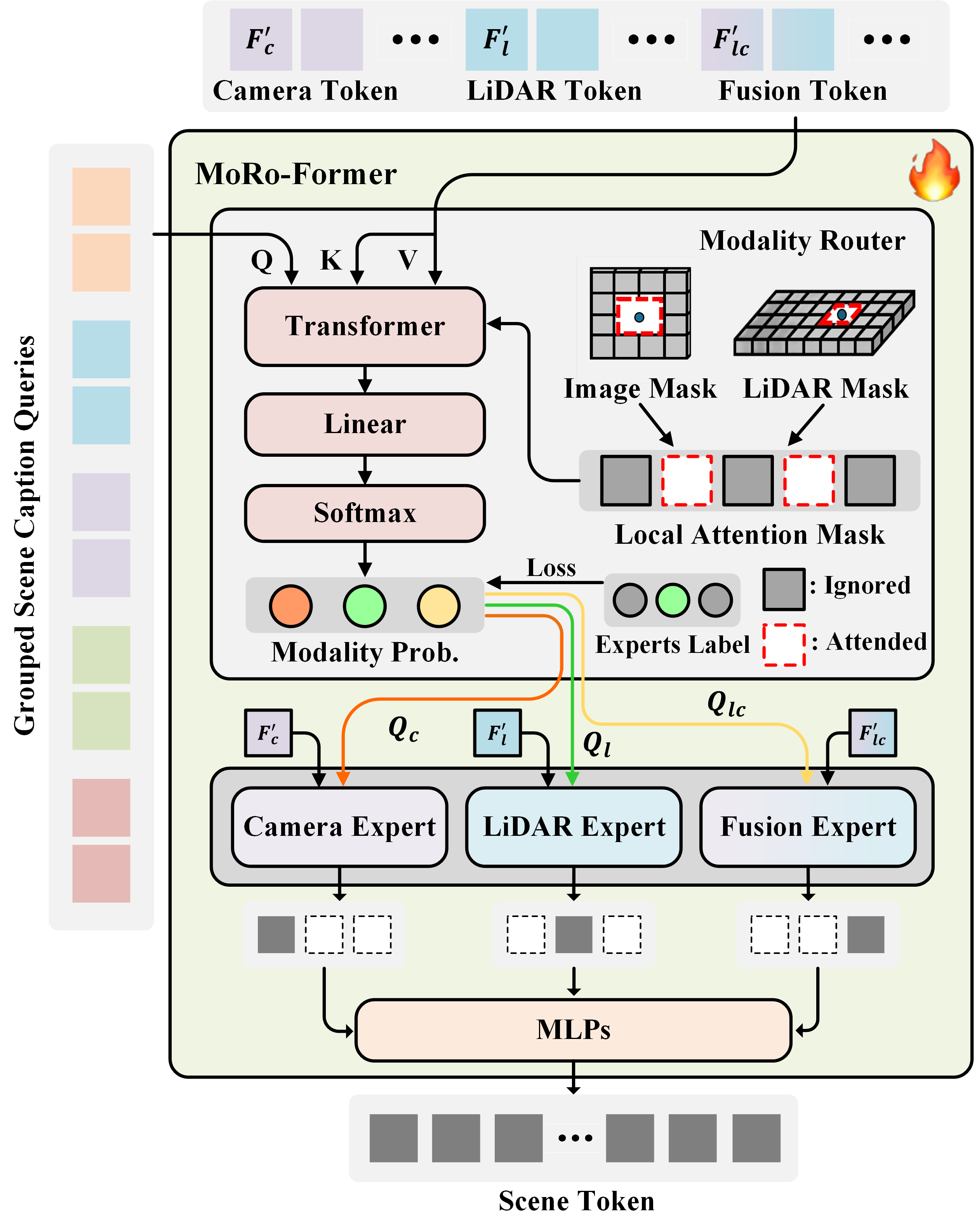}
    \caption{The MoRo-Former module. It performs locality-aware masked attention to predict modality probabilities and hard-route task queries to LiDAR, camera, or fusion experts. The routed features are aggregated and compressed into compact task tokens for the LLM.}
    \label{fig:moroformer}
\end{figure}

Prior bridging modules between multimodal encoders and LLMs mainly rely on simple MLP projections or standard Q-Former~\cite{met1} designs that treat all queries uniformly. Off-road driving, however, exhibits large condition-dependent sensor uncertainty (e.g., illumination and weather), and different tasks may benefit from different modalities. We therefore propose MoRo-Former (Fig.~\ref{fig:moroformer}), a locality-aware modality routing module that assigns task queries to modality experts and produces compact task tokens for the LLM.

We define $T{=}5$ tasks: weather description, drivable area, terrain traversability, obstacle detection, and driving suggestion. For each task, we allocate $K{=}64$ learnable queries, forming $\mathbf{Q}\in\mathbb{R}^{(TK)\times C_q}$. To mitigate inter-task interference, we inject task identity via a learnable group embedding $\mathbf{E}_g\in\mathbb{R}^{T\times C_q}$:
\begin{equation}
\tilde{\mathbf{q}}_{t,k}=\mathbf{q}_{t,k}+\mathbf{E}_g[t].
\label{eq:group_embed}
\end{equation}

For tasks requiring modality selection (traversability, obstacle detection, and driving suggestion), each query $\tilde{\mathbf{q}}_i$ is associated with a 3D reference point $\mathbf{r}_i=(x_i,y_i,z_i)^{\mathsf{T}}$, projected onto the LiDAR BEV and camera feature planes as $\bm{\pi}_l(\mathbf{r}_i)$ and $\bm{\pi}_c(\mathbf{r}_i)$. We construct a binary local mask $\mathbf{M}_i\in\{0,1\}^{H_lW_l+H_cW_c}$ over the flattened fused token sequence to attend only to tokens within fixed windows around these projections:
\begin{equation}
\begin{aligned}
\mathbf{M}_i(u)=\;&
\mathbb{I}\!\left[\| \bm{\phi}_l(u)-\bm{\pi}_l(\mathbf{r}_i)\|_{\infty}\le \frac{l_l}{2}\right] \\
&\vee\ 
\mathbb{I}\!\left[\| \bm{\phi}_c(u)-\bm{\pi}_c(\mathbf{r}_i)\|_{\infty}\le \frac{l_c}{2}\right],
\end{aligned}
\label{eq:local_mask}
\end{equation}
where $\bm{\phi}_l(\cdot)$ and $\bm{\phi}_c(\cdot)$ map token indices to LiDAR BEV and camera grid coordinates, respectively; $l_l$ and $l_c$ denote window sizes; and $\|\cdot\|_{\infty}$ defines a square neighborhood. Using masked cross-attention, a modality router predicts selection probabilities $\mathbf{p}_i=[p_{i,l},p_{i,c},p_{i,lc}]$ and performs hard assignment during inference:
\begin{equation}
m_i=\arg\max(\mathbf{p}_i),\quad m_i\in\{l,c,lc\}.
\label{eq:expert_assign}
\end{equation}

For each task $D$, we partition its queries into $\mathbf{Q}_D^{l}$, $\mathbf{Q}_D^{c}$, and $\mathbf{Q}_D^{lc}$ and decode them with the corresponding LiDAR/Camera/Fusion experts via cross-attention:
\begin{equation}
\mathbf{Q}_D^{\prime m}=\mathrm{CrossAttn}\!\left(\mathbf{Q}_D^{m},\,\mathbf{F}'_{m}\right),\quad m\in\{l,c,lc\}.
\label{eq:simple_crossattn}
\end{equation}

We then aggregate outputs with hard routing (each query keeps only its assigned expert output) and compress intra-task tokens within each branch to obtain compact expert tokens $\mathbf{S}_D^{l}$, $\mathbf{S}_D^{c}$, and $\mathbf{S}_D^{lc}$. The final multi-expert task representation is
\begin{equation}
\mathbf{S}_D=[\mathbf{S}_D^{l};\mathbf{S}_D^{c};\mathbf{S}_D^{lc}],\qquad
\hat{\mathbf{S}}_D=\mathrm{MLPs}(\mathbf{S}_D),
\label{eq:token_concat_mlp}
\end{equation}
which is concatenated with the prompt as the LLM context to generate scene descriptions and structured task outputs.

\subsection{Large Language Model and Path Planning} 

Large language models (LLMs) are crucial in the Wild-Drive framework, as high-quality scene descriptions are necessary to guide the path planner in generating reasonable trajectories for off-road scenarios.  Considering the computational limitations of field robots, Wild-Drive adopts efficient LLM backbones, including Qwen2.5-0.5B and Qwen2.5-3B~\cite{in8}, depending on the desired trade-off between efficiency and capability.

Specifically, user instructions are first tokenized into language tokens $\mathbf{x}_q \in \mathbb{R}^{L \times C}$, where $L$ is the token length and $C$ is the LLM's dimensionality. Next, the compact task tokens generated by MoRo-Former for the five Wild-Drive tasks are concatenated with the language tokens and input into the LLM. Inspired by \cite{in9}, we design a path planning Q\&A template with a special planning token $p$ to accumulate the LLM's understanding and reasoning context for the five tasks, expressed as:

 \begin{equation}
 \mathbf{s}\sim p_{\theta}\!\left(\mathbf{s}\ \middle|\ \mathbf{X}_q,\ \hat{\mathbf{S}}_{\mathcal{D}},\ \mathbf{x}_p\right),
 \label{eq:wilddrive_finalQA}
 \end{equation}
where $\mathbf{s}$ represents the LLM-generated answer, including the scene description and structured outputs for the five Wild-Drive tasks. $\mathbf{X}q$ is the tokenized user instruction sequence, and $\hat{\mathbf{S}}_{\mathcal{D}}$ is the concatenated representation of the MoRo-Former task tokens. $\mathbf{x}_{p}$ is the embedding of the special planning token $p$, used as a conditioning signal for downstream trajectory or action generation.

Finally, to bridge the gap between the LLM's language space and the continuous trajectory space, inspired by GenAD \cite{met2}, we design a gated recurrent unit (GRU) to decode the trajectory from the planning token $\mathbf{x}_{p}$:

\begin{equation}
\mathbf{y}_t = \mathrm{GRU}\left(\mathbf{x}_{p}, \mathbf{y}_{t-1}; \theta_{\text{GRU}}\right),
\label{eq:trajectory_gru}
\end{equation}

where $\mathbf{y}t$ represents the decoded trajectory point at time step $t$. $\mathbf{x}{_p}$ is the embedding of the planning token, serving as the conditioning signal for trajectory generation. $\mathbf{y}_{t-1}$ is the previous trajectory point, $\theta_{\text{GRU}}$ denotes the learnable parameters of the GRU. The GRU network uses the current planning token and the previous trajectory point to generate the next trajectory point. After obtaining the trajectory points, mature interpolation algorithms can be used to generate a continuous motion trajectory.

\subsection{Loss Functions}







To train Wild-Drive, it is crucial to guide the model to use sensor experts that are unaffected by the environment for decoding. To achieve this, we randomly drop camera or LiDAR inputs to encourage MoRo-Former to select the most reliable sensor expert for decoding. The drop probability is set to $1/3$, balancing three input settings: camera-only, LiDAR-only, and dual-modality. This strategy ensures that MoRo-Former can perform correct routing even in scenarios with corrupted sensors.

The training of MoRo-Former consists of two phases. In the first phase, all queries for five tasks are processed in parallel by each expert decoder and stacked, without dropping any sensors, directly proceeding to the subsequent scene captioning and path planning tasks. For LLM-generated templated natural language tasks, we use autoregressive cross-entropy loss \( L_{\text{ce}} \), and for paths regressed by the GRU decoder, we use L2 loss. The overall loss function \( L_{\text{total}} \) is composed of the following components:

\begin{equation}
L_{\text{total}} = L_{\text{text}} + L_{\text{waypoint}},
\end{equation}
where \( L_{\text{text}} \) is the text loss that encompasses the following five structured template-based Q\&A tasks: weather description, passable area, terrain feasibility, obstacle detection, and driving suggestions. The \( L_{\text{waypoint}} \) term represents the trajectory regression loss, which calculates the difference between the generated trajectory and the pre-estimated pose.



In the second phase, we freeze the pretrained expert decoders and train only the routing function in MoRo-Former. To simulate sensor corruption scenarios, we randomly drop either the camera or LiDAR input. The supervision label $y_i$ indicates the available modality: $[0,1,0]$ for camera-only, $[1,0,0]$ for LiDAR-only, and $[0,0,1]$ for dual-modality input. MoRo-Former is optimized with the cross-entropy loss \( L_{route} \):
\begin{equation}
L_{route} = \sum_{i=1}^{N} \mathrm{CE}(y_i, p_i),
\end{equation}
where $\mathrm{CE}$ measures the discrepancy between the routing probability $p_i$ and the target expert label $y_i$.

\section{The OR-C2P Benchmark}
We build the Off-Road Caption-to-Plan benchmark (OR-C2P) based on the ORAD-3D dataset~\cite{rw13}. The configuration details of the ORAD-3D dataset are provided in Section~V.A.

\begin{table}[t]
\centering
\small
\setlength{\tabcolsep}{0pt} 
\renewcommand{\arraystretch}{1.15} 
\caption{Detailed Data Types and Data Split of the OR-C2P Dataset}
\label{tab:orc2p} 
\begin{tabular*}{\columnwidth}{@{\extracolsep{\fill}}ccc@{}}
\toprule
\textbf{Dataset Name} & \textbf{OR-C2P} & \\ \midrule
\textbf{Weather Types} & 5 & \\
\textbf{Illumination Types} & 4 & \\
\textbf{Terrain Types} & 10 & \\
\textbf{Frames with Obstacles} & 19,527 & \\
\textbf{Train Sequences / Frames} & 100 / 39,727 & \\
\textbf{Validation Sequences / Frames} & 15 / 5,717 & \\
\textbf{Test Sequences / Frames} & 29 / 12,164 & \\ 
\textbf{Total Sequences / Frames} & 144 / 57,808 & \\ \bottomrule
\end{tabular*}
\end{table}

\subsection{LLM Q\&A Generation}


To fit the limited capacity of the lightweight LLM in Wild-Drive, we design concise, structured Q\&A templates for off-road scene understanding and planning (Tab.~\ref{tab:orc2p}), where each task uses a predefined answer space for stable training and evaluation. Supervision is generated by applying Qwen2.5-VL-72B to the collected off-road data and post-processing the outputs into canonical labels. Weather and terrain traversability are directly annotated with camera-based pseudo-labels, while drivable area and obstacle detection are first inferred from images and then refined with calibrated LiDAR point clouds through clustering and filtering. A final round of random human verification is conducted to ensure annotation quality.

\textbf{Weather:}
We represent weather as a paired label of \emph{weather} and \emph{illumination}. The model selects one item from each fixed vocabulary (e.g., weather: \texttt{\{sunny, cloudy, rainy, snowy, foggy\}}, illumination: \texttt{\{bright Light, daylight, twilight, darkness\}}), yielding a compact scene-caption cue for downstream reasoning.

\textbf{Drivable:}
We cast drivable area reasoning as a small closed-set classification problem. The LLM predicts (i) availability (e.g., \texttt{\{clear, partially\_blocked, blocked\}}) and (ii) a coarse free-space direction relative to the ego vehicle (e.g., \texttt{\{front, front\_left, front\_right, left, right\}}), avoiding dense segmentation while preserving planning-critical semantics.

\textbf{Terrain traversability:}
We ask the LLM to classify the near-field terrain and its traversability level using fixed vocabularies (e.g., terrain: \texttt{\{dirt, gravel, grass, mud, sand, snow, rock\}}; difficulty: \texttt{\{easy, moderate, hard, impassable\}}), explicitly summarizing traction and risk for the planner.

\textbf{Obstacle detection:}
We use a closed-set obstacle taxonomy and coarse localization. The LLM reports obstacle existence with a category label (e.g., \texttt{\{vehicle, pedestrian, animal, rock, tree, pole, building, unknown\}}) and a discretized position descriptor (direction: \texttt{\{front, front\_left, front\_right, left, right\}}; distance: \texttt{\{near, mid, far\}}), forming a structured tuple that can be directly consumed by planning.

\begin{figure}
    \centering
    \includegraphics[width=1\linewidth]{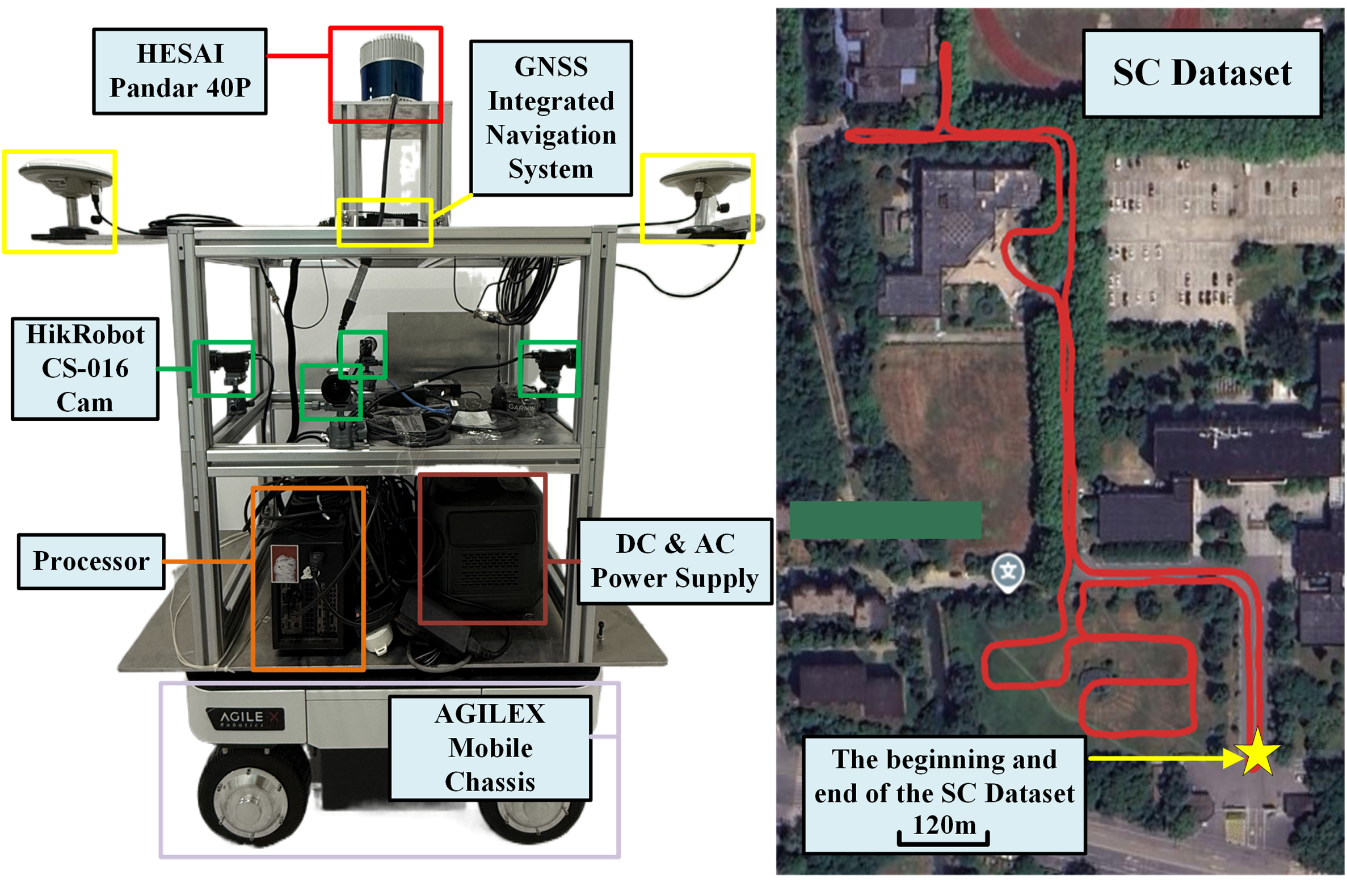}
    \caption{Experimental setup and data collection route map of the self-collected dataset}
    \label{fig:exper_setup}
\end{figure}

\begin{figure*}
    \centering
    \includegraphics[width=1\linewidth]{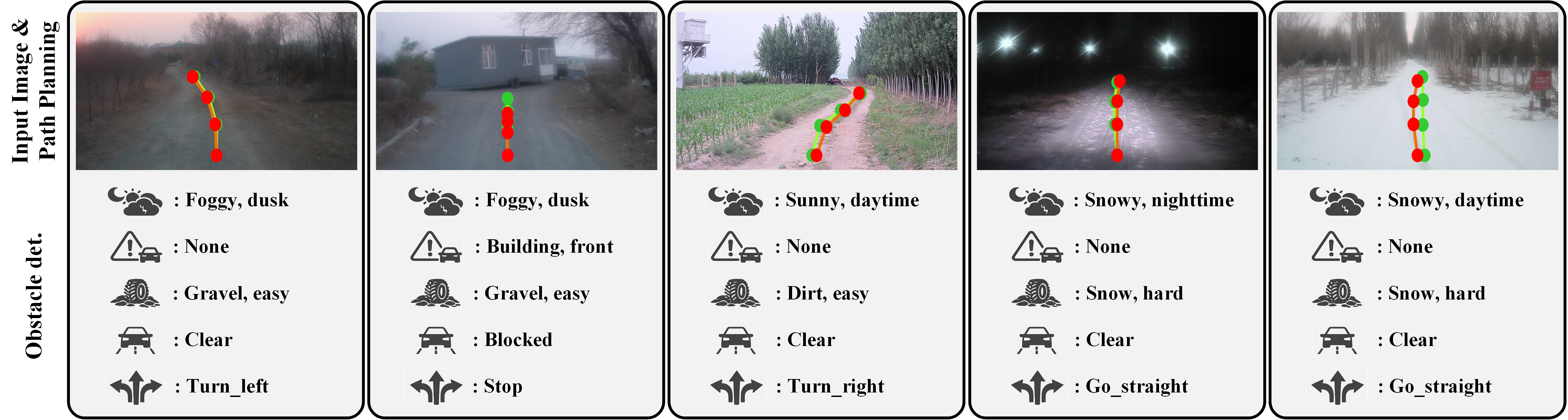}
    \caption{Quantitative analysis of Wild-Drive for scene captioning and path planning on the OR-C2P dataset. In the path planning visualizations, red indicates the model-predicted trajectory and green indicates the ground truth trajectory.}
    \label{fig:quan_vis}
\end{figure*}

\begin{table*}
\centering 
\small 
\setlength{\tabcolsep}{0pt} 
\renewcommand{\arraystretch}{1.2} 
\caption{Comprehensive Evaluation of Off-Road Scene Captioning Performance on the OR-C2P and SC Datasets (B-n represents BLEU-n; BERT-P represents BERT performance).} 
\label{tab:overall_results} 
\begin{tabular*}{\textwidth}{@{\extracolsep{\fill}}lcccccccc} 
\toprule 
\multirow{2}{*}{\textbf{Approach}} & \multirow{2}{*}{\textbf{Modality}} & \multicolumn{4}{c}{\textbf{OR-C2P Dataset}} & \multicolumn{2}{c}{\textbf{SC Dataset}} & \multirow{2}{*}{\textbf{Params}} \\ 
\cmidrule(lr){3-6} 
\cmidrule(lr){7-8} 
& & \textbf{B-1} ↑ & \textbf{B-2} ↑ & \textbf{B-4} ↑ & \textbf{BERT-P} ↑ & \textbf{B-1} ↑ & \textbf{BERT-P} ↑ & \\ 
\midrule 
Mini-GPT4~\cite{ex4} & V   & 31.27 & 16.57 & 5.52 & 77.43 & 29.52 & 74.26 & 7B \\ 
LLaVA1.5~\cite{ex5} & V & 36.75 & 18.25 & 9.71 & 81.34 & 33.27 & 71.55 & 7.06B \\ 
Instruct-BLIP~\cite{met1} & V & 37.59 & 19.75 & 10.29 & 82.89 & 33.41 & 70.52 & 7.91B \\ 
LLama-AdapterV2~\cite{ex7} & V & 49.17 & 38.46 & 27.29 & 81.61 & 46.33 & 79.92 & 7.22B \\ 
LiDAR-LLM~\cite{in10} & L & 68.04 & 58.61 & 46.11 & 87.49 & 66.19 & 88.47 & 6.96B \\ 
BEV-LLM-1B~\cite{in11} & V-L & 65.72 & 53.29 & 45.73 & 86.29 & 59.07 & 87.32 & 1.35B \\ 
BEV-LLM-8B~\cite{in11} & V-L & 67.11 & 57.95 & 47.93 & 88.17 & 61.26 & 88.91 & 8.17B \\ 
\textbf{Wild-Drive-0.5B (Ours)} & V-L & 69.72 & 56.44 & 47.21 & 95.79 & 63.25 & 89.53 & \textbf{0.71B} \\ 
\textbf{Wild-Drive-3B (Ours)} & V-L & \textbf{71.72} & \textbf{60.79} & \textbf{49.26} & \textbf{98.13} & \textbf{65.45} & \textbf{93.64} & 3.25B \\ 
\bottomrule 
\end{tabular*} 
\end{table*}

\textbf{Driving suggestion:}
To supervise high-level driving suggestions, we project the recorded trajectory into the ego-centric frame and extract the future $10$-second trajectory segment for each sample using the pre-estimated pose. We then perform $K$-means clustering on these future trajectories to obtain a compact action vocabulary:
\begin{equation}
\min_{\{\boldsymbol{\mu}_k\}_{k=1}^{K},\,\{a_i\}_{i=1}^{N}}
\sum_{i=1}^{N}\left\|\mathbf{v}_i-\boldsymbol{\mu}_{a_i}\right\|_2^2,
\end{equation}
where $\mathbf{v}_i$ denotes the vectorized future trajectory of sample $i$ in the ego frame, $\boldsymbol{\mu}_k$ are cluster centers, and $a_i\in\{1,\dots,K\}$ is the assignment. We map the resulting clusters to four action categories:
\texttt{\{go\_straight, turn\_left, turn\_right, stop\}},
and prompt the LLM to select exactly one action given current observations and scene context.

Overall, these structured Q\&A templates convert complex off-road perception into a small set of canonical labels, stabilizing learning with an efficient LLM and providing interpretable intermediate outputs for trajectory generation.

\subsection{Path Planning}

In off-road environments, unreliable GPS signals often prevent autonomous vehicles and robots from obtaining accurate high-precision localization. To address this issue, we employ KISS-ICP~\cite{ben1} to accurately register LiDAR point clouds, yielding a continuous and consistent pose sequence for each data segment. To obtain smooth and temporally continuous future trajectory supervision, we further apply B-spline interpolation to the pose-based trajectory, generating continuous path points along with their associated timestamps. For the Wild-Drive path planning task, we take the trajectory points at future horizons $t=\{1,2,5,10\}$ seconds relative to the current timestamp as the planning ground truth.

\section{Experiments}

\subsection{Datasets and Metrics}

We conduct experiments on two datasets: the OR-C2P benchmark and our Self-Collected (SC) dataset. OR-C2P is built on ORAD-3D~\cite{rw13} and contains 57K aligned LiDAR-camera pairs with GNSS ground-truth data. It covers common off-road sensor degradation conditions, including rain, snow, fog, and low-light scenes, and consists of 145 sequences collected from diverse off-road environments across China from spring to winter. Each sequence spans about 100 meters, with RGB images at a resolution of 1280×720. For the SC dataset, we use an robotic platform equipped with a Hesai Pandar 40P LiDAR and a Hikrobot CS-016 camera (Fig.~\ref{fig:exper_setup}) to collect a continuous trajectory of over 4 km. The off-road segments are extracted for real-world zero-shot generalization evaluation.



For the scene captioning task, we evaluate Wild-Drive using standard NLP metrics. The BLEU score ~\cite{ex1} measures n-gram precision, where a score of 1 indicates identical text. BLEU-n (B-n) evaluates the precision of unigrams, bigrams, trigrams, and four-grams. The BERT-score performance (BERT-P)~\cite{ex2} assesses semantic similarity by comparing contextual embeddings, instead of literal word matches. Specifically, we evaluate the performance using BLEU-1, BLEU-2, and BLEU-4 to capture both basic word overlap and more complex n-gram matches.

For the path planning task, we use the Final Displacement Error (FDE), computed as \( \lVert s_{T-1} - s^{*}_{T-1} \rVert_2^2 \), where \( s \) and \( s^{*} \) represent the predicted and ground-truth trajectories. FDE evaluates the accuracy of the final predicted waypoint, which serves as the goal for local planning (e.g., obstacle avoidance). For multimodal trajectory predictors, we use the minimum average displacement error (minADE).

\subsection{Implementation Details}

We implement two Wild-Drive variants with different LLM backbones: Wild-Drive-0.5B with Qwen2.5-0.5B-Instruct and Wild-Drive-3B with Qwen2.5-3B-Instruct, containing 0.5B and 3B parameters, respectively. Both variants are fine-tuned for instruction-following tasks. MoRo-Former employs 320 learnable queries, with 64 queries assigned to each task. For visual encoding, we use DINOv3 ViT-S/16 distilled (21M) and freeze all its layers during fine-tuning. All experiments are conducted on a workstation with an Intel Core i9-12900K CPU and an NVIDIA RTX 4090 GPU. The Qwen models are fine-tuned using LoRA~\cite{ex3} with rank and alpha both set to 16.

\subsection{Performance Analysis}

\paragraph{Qualitative results}
We present qualitative results of Wild-Drive in Fig.~\ref{fig:quan_vis}. Despite using an efficient large language model, Wild-Drive-0.5B captures key off-road cues (e.g., illumination/weather, terrain, drivable free space, and obstacles) and generates reasonable future trajectories under diverse conditions. The scene captions are structured and task-oriented, which helps the planner behave consistently in unstructured environments.

Compared with BEV-LLM~\cite{in11}, Wild-Drive benefits from MoRo-Former, which performs routing-based cross-modal aggregation. When one modality is degraded, the router emphasizes the most reliable expert. For instance, under low-light/nighttime conditions, traversability and obstacle queries are more often routed to the LiDAR expert, yielding more stable obstacle localization and safer trajectories. When LiDAR becomes sparse or corrupted, queries shift to the camera or fusion expert to leverage visual cues. This routing strategy improves the robustness of structured Q\&A outputs and mitigates failures from indiscriminate multimodal fusion.



\begin{table}[t]
\centering
\small 
\caption{Comprehensive Evaluation of Off-Road Path Planning Performance on the OR-C2P Datasets.}
\begin{tabular}{llcc}
\toprule
\textbf{Approach} & \textbf{Method Type} & \textbf{FDE $\downarrow$} & \textbf{minADE $\downarrow$} \\
\midrule
CoverNet~\cite{ex8} & Classif.-based & 2.47 & 1.31 \\
MTP~\cite{ex9} & Reg.-based & 1.59 & 0.84 \\
MultiPath~\cite{ex10} & Anchor-based & 1.56 & 0.92 \\
TopoPath~\cite{rw10} & Transf.-based & \textbf{0.92} & \textbf{0.43} \\
\midrule
BEV-LLM-GRU~\cite{in11} & LLM-based & 1.33 & 0.97 \\
\textbf{Wild-Drive (Ours)} & LLM-based & 1.09 & 0.66 \\
\bottomrule
\end{tabular}
\label{tab:approaches_comparison}
\end{table}

\paragraph{Quantitative results}
We quantitatively evaluate Wild-Drive on the OR-C2P benchmark and compare it with state-of-the-art methods for scene captioning and path planning (Tab.~\ref{tab:overall_results}). Specially due to the substantial reliance of the ${\mathrm{TOD}}^{3}\mathrm{Cap}$ method on 3D annotations, we have excluded it from our comparison. As LIDAR-LLM is not open-source, we have replicated and trained it for validation purposes.

Scene Captioning: 
We evaluate Wild-Drive on the OR-C2P and SC datasets and compare it with previous off-road scene captioning methods~\cite{ex4, ex5, met1, ex7, in10, in11} (Tab.~\ref{tab:overall_results}).
On the OR-C2P dataset, Wild-Drive-3B achieves the best performance across BLEU-1/2/4 and BERT-P, with BLEU-4 of 49.26 and BLEU-1 of 71.72, outperforming BEV-LLM-8B and LiDAR-LLM, respectively. It also achieves the highest semantic similarity, with BERT-P reaching 98.13, surpassing BEV-LLM-8B.
Wild-Drive-0.5B, with only 0.71B parameters, remains competitive, achieving BLEU-4 of 47.21 and a higher BLEU-1 than LiDAR-LLM (69.72 vs.~68.04), despite having fewer parameters.
On the SC dataset, Wild-Drive-3B ranks first in BERT-P (93.64), though slightly lower in BLEU-1 than LiDAR-LLM (65.45 vs.~66.19), indicating improved semantic alignment.
Finally, Wild-Drive achieves these results with fewer task queries (320) than BEV-LLM (512) and LiDAR-LLM (576), highlighting the efficiency of MoRo-Former’s routing and compression.

Path Planning:
We evaluate Wild-Drive on the OR-C2P dataset and compare it with prior off-road path-planning methods (Tab.~\ref{tab:approaches_comparison}). For fair comparison, we equip BEV-LLM-1B with the same GRU trajectory decoder and retrain it under identical settings, denoted as BEV-LLM-GRU. Wild-Drive achieves an FDE of 1.09 and a minADE of 0.66, outperforming BEV-LLM-GRU by 0.24 and 0.31, respectively, and ranking best among LLM-based methods. Compared with traditional methods, it also yields lower endpoint and average displacement errors, highlighting the advantage of language-conditioned multimodal understanding over purely geometric or anchor-based approaches. Wild-Drive still underperforms the specialized planner TopoPath by 0.17 in FDE and 0.23 in minADE, indicating the advantage of dedicated planning models. Overall, these results demonstrate the competitiveness of our perception-to-planning framework, while suggesting that further gains may require tighter integration of motion modeling and language-conditioned representations.

\begin{table}[t]
\centering
\small
\caption{Ablation study results for Wild-Drive evaluating the contribution of each key component on the OR-C2P dataset.}
\begin{tabular}{lccc}
\toprule
\textbf{Approach} & \textbf{B-1 $\uparrow$} & \textbf{B-4 $\uparrow$} & \textbf{BERT-P $\uparrow$} \\
\midrule
\textbf{Wild-Drive (Ours)}        & \textbf{71.72} & \textbf{49.26} & \textbf{98.13} \\
\midrule
w/ DINOv3 \& Q-F         & 63.41 & 44.59 & 86.24 \\
w/ V.Net \& Q-F          & 60.39 & 41.72 & 83.51 \\
w/ DINOv3 \& V.Net \& Q-F            & 66.79 & 45.92 & 92.74 \\
\bottomrule
\end{tabular}
\label{tab:ablation}
\end{table}



\subsection{Ablation Study}

We conduct ablations on the OR-C2P benchmark to analyze the contribution of each component to scene captioning (Tab.~\ref{tab:ablation}). Removing either the camera or LiDAR branch results in a clear drop in BLEU and BERT-score, showing that off-road captioning benefits from multimodal complementarity: the camera provides appearance cues, while LiDAR offers stable geometric evidence under visual degradations. Replacing MoRo-Former with a standard Q-Former using the same number of queries further degrades BLEU and BERT-score by about $5.4\%$, yielding performance close to BEV-LLM. These results indicate that uniform query-based bridging is insufficient under modality-dependent uncertainty, whereas routing and compression enable more reliable expert aggregation for structured captions.

\subsection{Runtime}
We report the runtime of Wild-Drive under two LLM configurations to demonstrate its low computational cost and plug-and-play applicability. All measurements are conducted on a workstation equipped with an Intel Core i9-12900K CPU and an NVIDIA RTX 4090 GPU. Wild-Drive-0.5B runs at an average latency of 1.271,s per sample, which is reduced to 0.542,s with 4-bit quantization. Wild-Drive-3B incurs 4.075,s on average, and achieves 1.653,s with 4-bit quantization. We also evaluate the system on an in-vehicle computer equipped with an NVIDIA RTX 2080Ti GPU, where the average runtime under the non-quantized setting is 37\% longer than that on the NVIDIA RTX 4090 GPU. These results indicate that Wild-Drive can be deployed with reasonable latency, and quantization further improves efficiency without changing the overall pipeline.

\section{CONCLUSION}

This paper presents Wild-Drive, a unified framework for off-road scene captioning and path planning, designed for interpretable understanding and reliable planning under frequent sensor corruptions in unstructured environments. Wild-Drive extracts camera and LiDAR features with strong backbones, and uses MoRo-Former for task-conditioned modality routing and token compression to adaptively fuse reliable information under modality degradation. It further integrates an efficient LLM, a planning token, and a GRU decoder to generate structured captions and predict future trajectories. We also introduce the OR-C2P benchmark for structured caption-to-planning evaluation in degraded off-road scenarios. Experiments on OR-C2P and self-collected datasets show that Wild-Drive outperforms prior LLM-based methods with better stability, while ablation and runtime studies confirm the effectiveness and real-time potential of its key components. We hope Wild-Drive and OR-C2P provide a reproducible foundation for future research on explainable off-road autonomous driving.




\end{document}